\RequirePackage{fix-cm}
\documentclass[smallextended]{svjour3}       
\smartqed  
\usepackage{graphicx}
\usepackage{subfigure}
\begin{document}

\title{Underwater image enhancement with Image Colorfulness Measure
}


\author{Hui Li         \and
        Xi Yang  \and
        ZhenMing Li \and
        TianLun Zhang
}



\date{Received: date / Accepted: date}

\maketitle

\begin{abstract}
Due to the absorption and scattering effects of the water, underwater images tend to suffer from many severe problems, such as low contrast, grayed out colors and blurring content. To improve the visual quality of underwater images, we proposed a novel enhancement model, which is a trainable end-to-end neural model. Two parts constitute the overall model. The first one is a non- parameter layer for the preliminary color correction, then the second part is consisted of parametric layers for a self-adaptive refinement, namely the channel-wise linear shift. For better details, contrast and colorfulness, this enhancement network is jointly optimized by the pixel-level and characteristic-level training criteria. Through extensive experiments on natural underwater scenes, we show that the proposed method can get high quality enhancement results.
\keywords{First keyword \and Second keyword \and More}
\end{abstract}

\section{Introduction}
\label{intro}
With the significant growth of attention devoted into the exploration in ocean, river and lake, underwater imaging has emerged as a relevant research in wide-ranging fields, such as autonomous underwater vehicles \cite{leonard2016autonomous}, biological resource investigation \cite{melman2015distributed}, etc. The major problems in underwater imaging processing are color cast and contrast degradation since the absorption and scattering effect in water \cite{MIP}. Both of these problems tend to limit the visibility of underwater objects. Therefore, it is important to improve the degraded quality of underwater images.

The focus of underwater image improvement is to restore colors and enhance contrast for high perceived quality. Obviously, this vision task is an inherently ill-posed problem as one clean underwater image can have connection with multifarious degraded scenarios. Moreover, when these adverse factors (e.g., blurring effects, low contrast and grayed out color) are more serious, this connection will become more ambiguous since fine details of the underwater background have little or no evidence in the corresponding version with reduced visibility. Therefore, in absence of rich additional images or information [????], the enhancement on single underwater image is an extremely challenging vision issue [????].

To make this issue well-posed, numerous worthwhile explorations have been conducted for single underwater image enhancing purpose. To correct shifted colors, white balancing \cite{liu1995automatic} is the most popular tool, meanwhile, some other illumination models are also investigated for color rendition \cite{UDCP,liu2016underwater,peng2018generalization}. In addition, several methods of single image haze removal are introduced to deal with blurring effect and low contrast in underwater images. In these methods, the authors pointed that the degradation process in underwater is similar to the atmospheric absorption and scattering \cite{UDCP}. Thus the imaging models about haze (such as the atmospheric scattering model\cite{narasimhan2002vision,narasimhan2000chromatic,fattal2008single,tan2008visibility}) are used to describe the formation of underwater images. And some successful de-hazing algorithms (e.g., dark channel modeling \cite{DCP} and polarization haze removal method \cite{schechner2003polarization}) are also incorporated into the family of underwater enhancement.

Recently, deep learning has been widely deployed in the field of image enhancement, such as haze removal \cite{yang2019single}, de-raining \cite{zhang2019image}, de-noising \cite{liu2018multi} and underwater image improvement \cite{GAN}. Deep learning is a typical data-driven method which can learn the distributive characteristics of data without strong assumptions and priors. Thus most image enhancement methods tend to use deep neural networks as the map from degraded images to the recovered results \cite{GAN}. Different from existing work using deep neural model to directly recover images, in this paper, we develop a connectionist model to perform the color correction and the channel-wise shift in an end-to-end manner. To this end, we use this deep model to predict the shifting weights and biases. More concretely, our work can be concluded as follows.

(1) We propose a novel neural model for underwater image enhancement. This connectionist structure is composed of two functional modules. The first one is a non-linear shift for color correction, then a channel-wise refinement is conducted by the second module whose mapping coefficients are learned by a convolutional neural network.

(2) We proposed a joint optimization model for the neural network. This model is a fusion of a supervised learning and a unsupervised learning. The supervised learning is used for the pixel-wise alignment between the prediction and the corresponding ground truth. The unsupervised learning is designed for the preserving image characteristics.

(3) We proposed an unsupervised loss based on underwater image colorfulness measure. It can be proved from experiments that loss can well solve the color-casting  problem of underwater images and the problem of color desaturation due to limited lighting conditions.

\section{Related Work}
\label{sec:1}

Image enhancement and restoration are fundamental topics in the field of computer vision. These topics contains many popular applications, such as de-hazing \cite{yang2019single}, rain removal \cite{zhang2019image}, super-resolution \cite{kim2016accurate}, underwater image enhancement \cite{GAN}, etc. Among these applications, the improvement of underwater images is a important researching filed for the development of marine programs, e.g., protecting coral reefs \cite{roelfsema2018coral}, deep ocean exploration and underwater teleoperation \cite{zhang2017force}.

Typically, the approaches to enhance underwater images fall into two umbrellas. One is based on multiple images or additional information. Yoav et al. \cite{schechner2007regularized} proposed an adaptive filtering method to the regularized image recovery in scattering media. They used different degrees of polarization images to the visibility enhancement of a underwater image. Krotkov et al. \cite{cozman1997depth} presented an analysis on static scene via capturing different images under multiple visibility scenes. Despite impressive results, this method requires a static camera and significant change of illumination condition. Kocak et al. \cite{kocak2008focus} provided an overview of countering the degradation in long-range visibility. Some additional techniques were highlighted in this work, such as structured and multi-source lighting. In addition, several methods that achieve specialized knowledge (e.g., scene depth and contrast \cite{chiang2011underwater,kaeli2011improving,carlevaris2010initial}) have also been introduced into underwater image processing.

Another is only based on single underwater image. In this case, the image enhancement becomes a under-constrained problem, which is more challenging than the first one mentioned above. To break down the limitation of underwater medium, Ancuti et al. \cite{ancuti2012enhancing} developed a fusion principle by deriving two characteristics and four weight maps from a single underwater image. Roser et al. \cite{roser2014simultaneous} proposed a joint method which can simultaneously conducts image quality assessment, enhancement and improved stereo. A similar method is introduced by Fu et al. \cite{fu2014retinex}. In this method, the reflectance and illumination are decomposed via a variational framework for Retinex, then, the reflectance and illumination are improved by different strategies of addressing the under-exposure and blurring issues. Zhang et al. \cite{zhang2017underwater} also employed Retinex model for the illumination adjustment of underwater image. Based on the piece-wise linear transformation, Fu et al. \cite{fu2017two} proposed an effective strategy to deal with the color distortion, and this method can improve the low contrast while reducing artifacts. Recently, deep learning models have drawn much attention and got many successful applications in different fields [????]. Fabbri et al. \cite{GAN} introduced a neural network into the enhancement of underwater images. The generative adversarial network is used in their work, in which the generative network takes underwater image as input and outputs enhanced result, through solving a minimax problem, the adversarial network plays a pivotal role to improve the performance of the generative network. Moreover, some popular methods of single image haze removal also are used in underwater image processing \cite{DCP}.

\section{The Proposed Work}
\begin{figure}[t]\centering  
	\includegraphics[width=\textwidth]{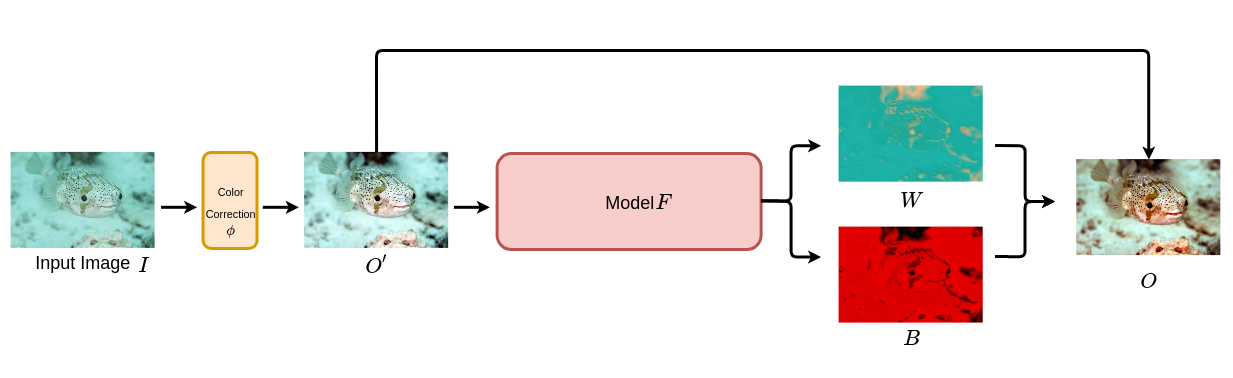}	
	\caption{Overview of the proposed underwater enhancement method}
	\label{fig:arch}
\end{figure}

In this section, we will introduce this proposed framework for the enhancement of underwater images. As illustrated in Figure \ref{fig:arch}, nonlinear mapping of the neural model output $W$,$B$ and the result of the color correction layer without parameters $O'$ to get the final underwater enhancement result $O$. More details will be described in what follows.

\subsection{Model Design}

The input underwater image and the corresponding enhanced result are denoted as $I$ and $O$, respectively. The first neural layer is designed for the preliminary color correction, and can be formulated as follows

\begin{equation}
O'=\phi(\theta(I^c(i),V_{min}^c,V_{max}^c),V_{min}^c,V_{max}^c), c \in \{R,G,B\}, i \in [1,N]
\end{equation}
in which $i$ represents the $i$th pixel of each channel, $N$ is the number of pixels in each channel, $\theta$ is a bilateral restraint function:
\begin{equation}
\theta(I^c,V_{min}^c,V_{max}^c) = min(max(I^c,V_{min}^c),V_{max}^c)=I^{\theta},
\end{equation}
here, first and 99th percentiles of pixel value distribution for each channel are used as $V_{min}^c$ and $V_{max}^c$. $\theta$ uploads the pixel values before the non-linear activation function $\phi$ that is an affine transform and can be defined as 

\begin{equation}
\phi(I^{\theta},V_{min},V_{max})=\frac{I^{\theta}-V_{min}}{V_{max}-V_{min}}.
\end{equation}

\begin{figure*}
	\centering
	\begin{minipage}{0.25\textwidth}
		\centering
		\includegraphics[width=\textwidth]{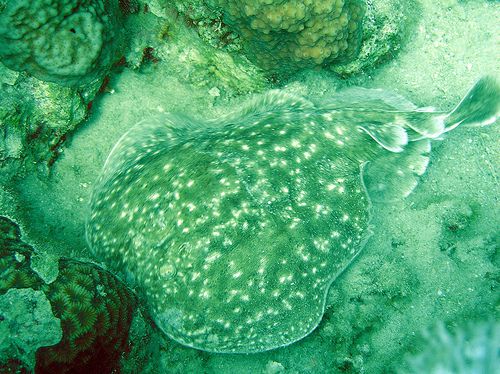}
	\end{minipage}
	\begin{minipage}{0.25\textwidth}
		\centering
		\includegraphics[width=\textwidth]{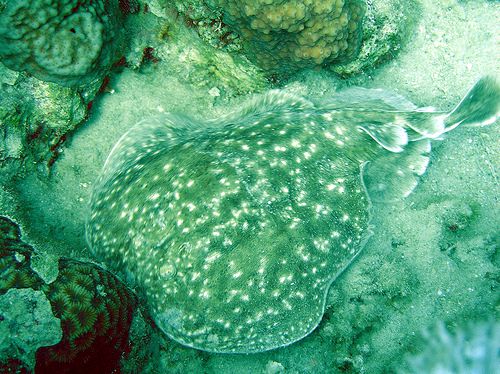}
	\end{minipage}
	\begin{minipage}{0.25\textwidth}
		\centering
		\includegraphics[width=\textwidth]{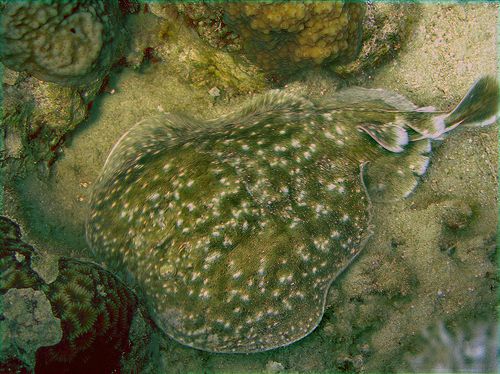}
	\end{minipage}
	\begin{minipage}{0.25\textwidth}
		\centering
		\includegraphics[width=\textwidth]{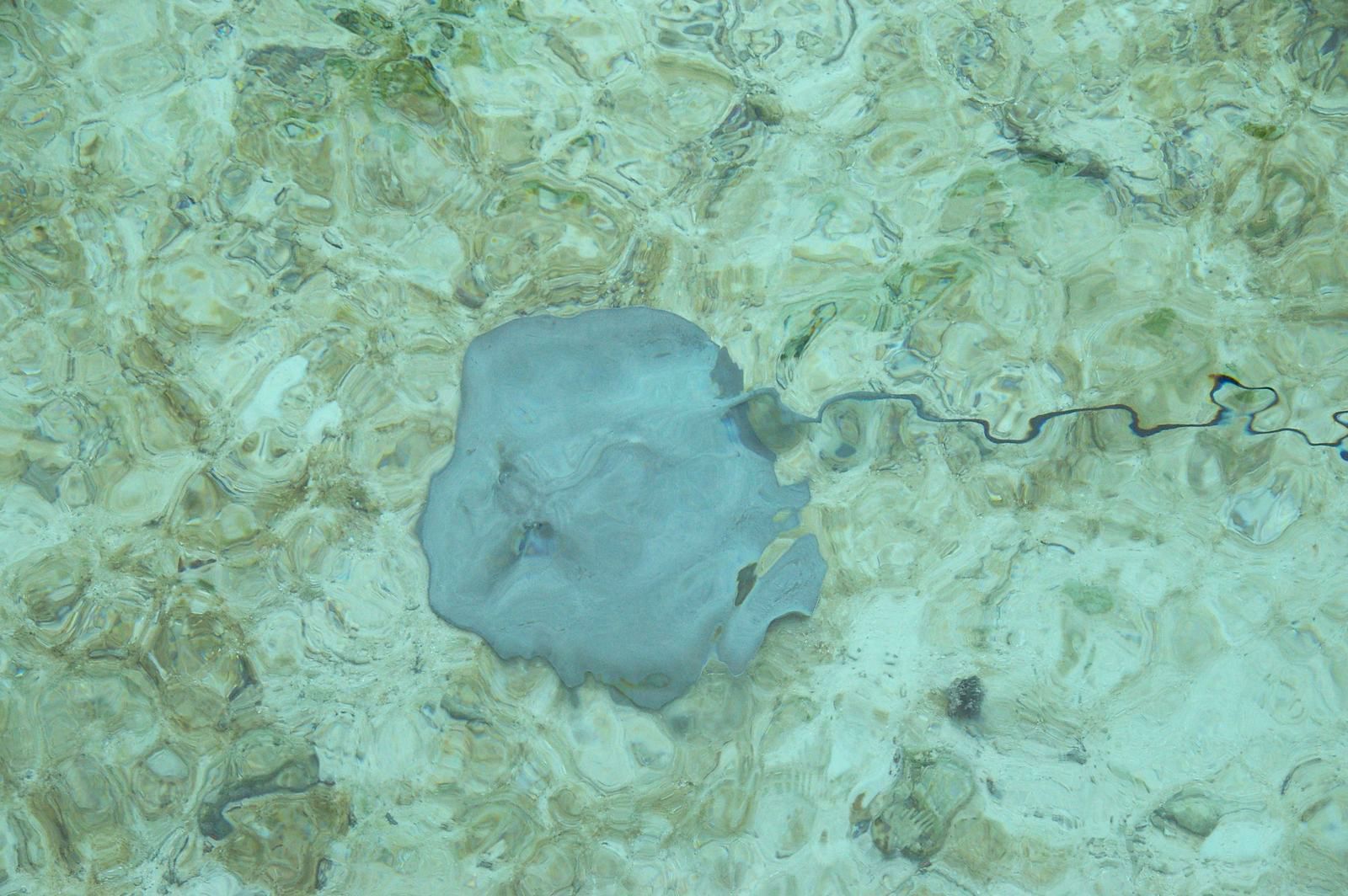}
	\end{minipage}
	\begin{minipage}{0.25\textwidth}
		\centering
		\includegraphics[width=\textwidth]{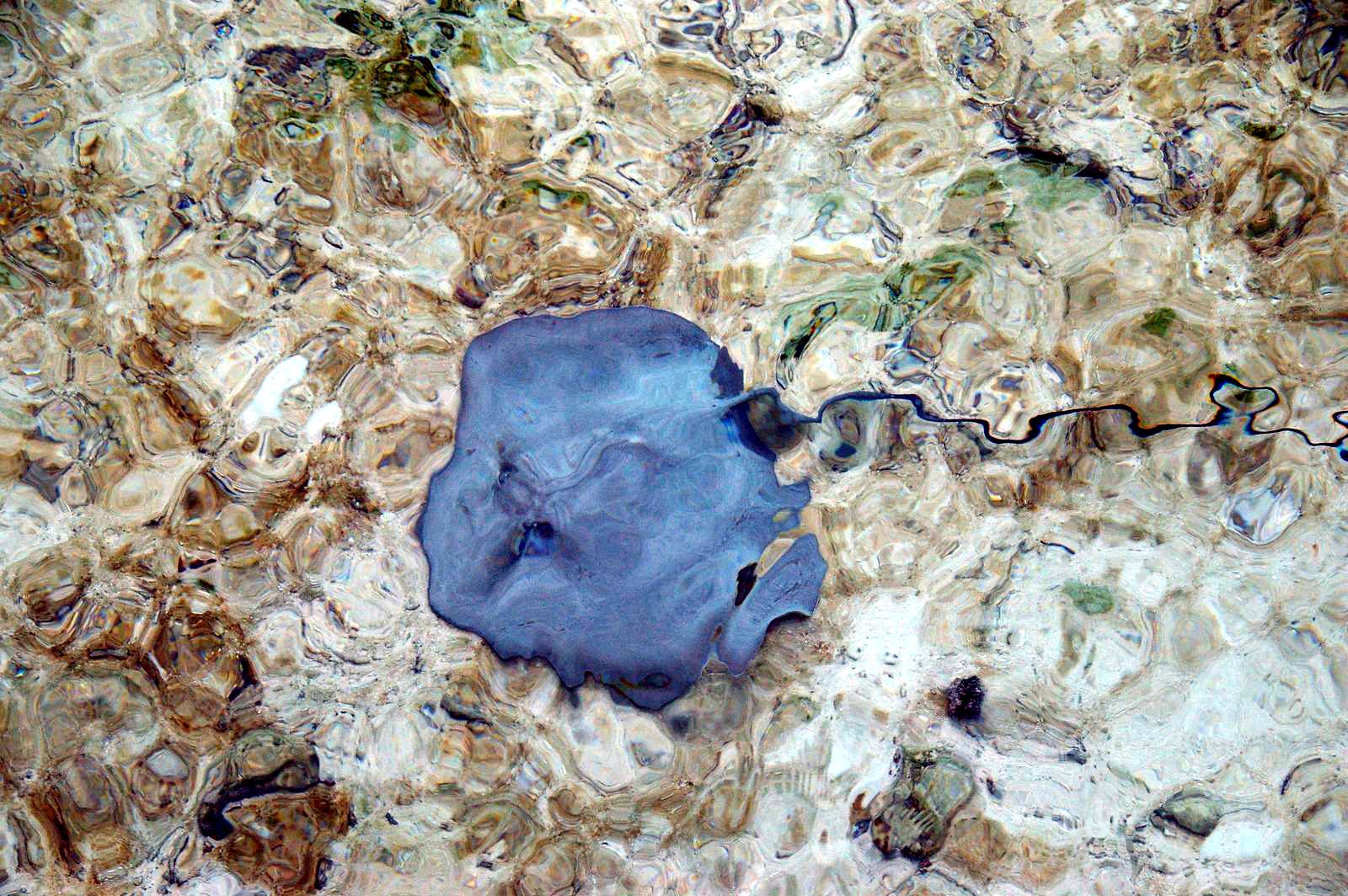}
	\end{minipage}
	\begin{minipage}{0.25\textwidth}
		\centering
		\includegraphics[width=\textwidth]{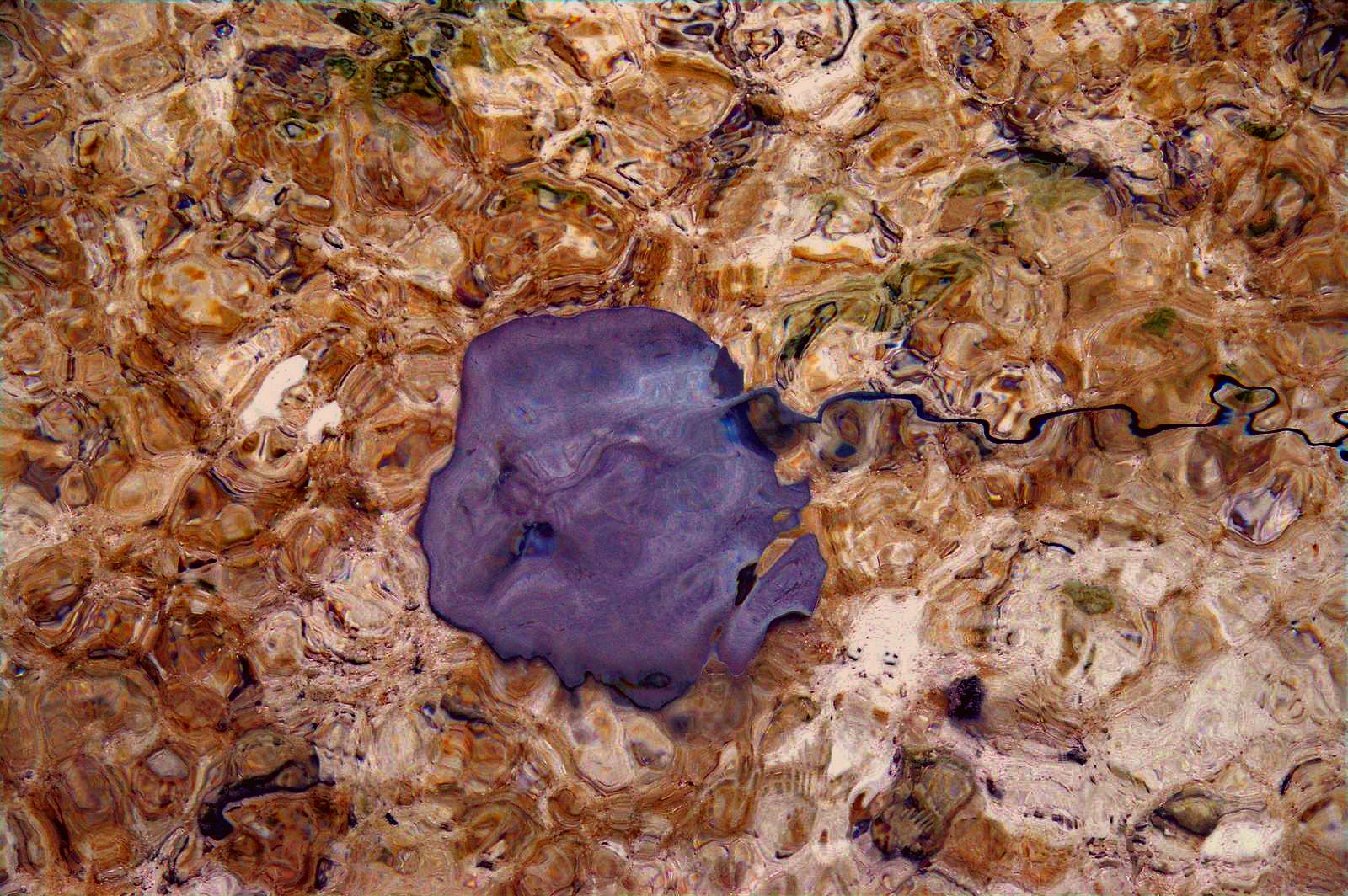}
	\end{minipage}
	
	\begin{minipage}{0.25\textwidth}
		\centering
		\includegraphics[width=\textwidth]{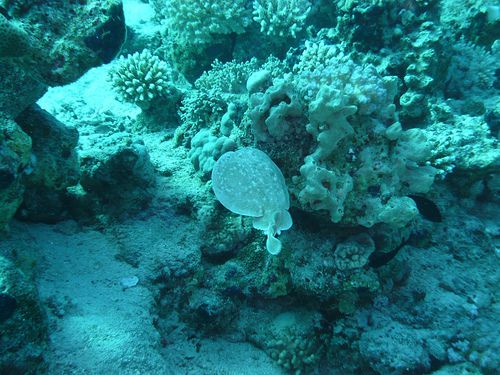}
		\subfigure{$I$}
	\end{minipage}
	\begin{minipage}{0.25\textwidth}
		\centering
		\includegraphics[width=\textwidth]{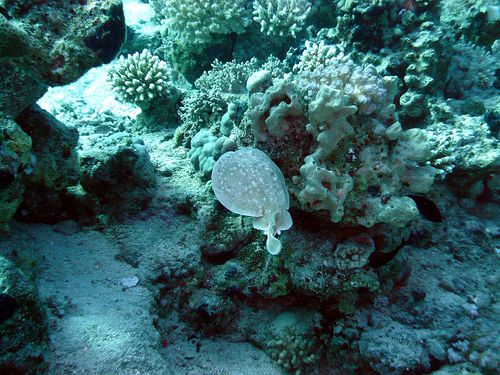}
		\subfigure{$O'$}
	\end{minipage}
	\begin{minipage}{0.25\textwidth}
		\centering
		\includegraphics[width=\textwidth]{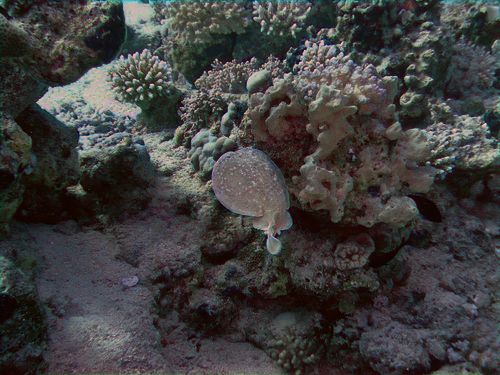}
		\subfigure{$O$}
	\end{minipage}
	
	\caption{Comparison of the color correction layer and the final output image of the model.}
	\label{fig:wb}
\end{figure*}

As shown in Figure \ref{fig:wb}, the resulting images derived from the preliminary color correction have been enhanced, but under different underwater lighting conditions, good or bad enhancement results will be obtained. So a further improvement is necessary for favorable visual quality. Thus we conduct a channel-wise no-linear shift in the following step. To this end, we model the model output $O''$ as 

\begin{equation}
O''=ReLU(W\phi(I^{\theta},V_{min},V_{max})+B)
\end{equation}
where $W$ is the weight factor matrix of the linear shift in the channel wise, $B$ is the bias matrix. Since connectionist model can explore and learn image characteristics in a flexibility manner, we develop a deep convolutional neural network with homogeneous kernels to predict $W$ and $B$ for different $I^{\theta}$. This adaptive method for non-linear shift can be described as follows

\begin{equation}
[W,B]=F(I^{\theta})
\end{equation}
where $F$ is the neural network model that is used to extract $W$ and $B$. In this paper, we use a network like VGG. The final output of the model is a 6-channel feature map that is the same size as the input image. The first 3 channels are used as $W$, and the last 3 are used as $B$.

To keep values of $O$ within the right range, a normalization operation is conducted as follows

\begin{equation}
O=\frac{O''}{max(O'')}
\end{equation}

$max(O'')$ represents the maximum value on each channel. The resulting enhancement results $O$ are shown in Fig. \ref{fig:wb}. It can be clearly seen that the result of the color correction layer is further modified and enhanced.

\subsection{Objective Function}

In this work, two types of objective function are used to optimize the neural model jointly. The first criterion is a pixel-wise alignment between $O$ and corresponding ground truth $\hat{O}$, namely,

\begin{equation}
L_{pixel}=\frac{1}{N}\sum^N_{i}(O(i)-\hat{O}(i))^2
\label{eq:pixel}
\end{equation}
where $i$ represents the $i$th pixel of each channel. $L_{pixel}$ can ensure that the output of the neural network is consistent with the label at pixel level, which means that the output of the network is the same as $O$.

Different from Eq. \ref{eq:pixel}, the second loss function is designed for unsupervised learning. Due to the effect of underwater lighting, the color information in underwater images is seriously lost. The human visual system(HVS) was introduced in the unsupervised loss to restore the human color perception of underwater images.

In \cite{hvs1,hvs2,hvs3}, HVS is more sensitive to relative changes than absolute changes in contrast. As shown in the following formula, two components are used in the underwater image colorfulness measure(UICM):
\begin{equation}
I^{RG}=I^R-I^G
\end{equation}
\begin{equation}
I^{YB}=\frac{I^R+I^G}{2}-I^B
\end{equation}

These two components in opponent-color theory are related to the chrominance of the image.

According to the color research of natural images \cite{hasler2003measuring,gao2013no}, the colorfulness in image can be expressed as a function of image statistical values. However, as mentioned above, underwater images will be seriously affected by lighting conditions, which will not only cause loss of color information, but also cause partial image details to be lost. Therefore, the image statistical values will be calculated using some pixel values of the image, just like the color correction layer mentioned earlier.

\begin{equation}
\mu_{RG}=\frac{\sum^{N-T_{max}}_{i=T_{min+1}}I^{RG'}(i)}{N-T_{min}-T_{max}}
\end{equation}
where $T_{min}$ is number of pixels with all pixel values below $V_{min}^{RG}$, while $T_{max}$ is the number of pixels with all pixels higher than $V_{max}^{RG}$. $I^{RG'}$ in the above formula represents $I^{RG}$ sorted by pixel value. $\mu_{YB}$ is also calculated in the same way. 

To ensure that there is not a certain color in the restored image (such as a large amount of green or blue in the underwater image), an intuitive observation method is that $(I^{RG}-I^{YB})$ approaches 0. And then, this measurement can be done by calculating $\mu_{RG}$ and $\mu_{YB}$.

Based on the $\mu_{RG}$ calculated previously, $\sigma_{\alpha,RG}$ can be calculated:
\begin{equation}
\sigma^2_{RG}=\frac{1}{N}\sum^N_{i}(I^{RG}(i)-\mu_{RG})^2
\label{eq:sigma}
\end{equation}

The variance can represent the dispersion of the value, that is, the larger the variance, the more discrete the value. Here the larger $\sigma_{\alpha,RG}$ and $\sigma_{\alpha,YB}$, the more diverse the colors.

Combining the two statistics mentioned above, $L'_{UICM}$ can be defined as:
\begin{equation}
L'_{UICM}=\alpha_1\sqrt{\mu_{\alpha,RG}^2+\mu_{\alpha,YB}^2}+\alpha_2\sqrt{\sigma_{\alpha,RG}^2+\sigma_{\alpha,YB}^2}
\label{eq:uicm'}
\end{equation}

In Eq. \ref{eq:uicm'}, coefficients $\alpha_1$ and $\alpha_2$ are obtained by linear regression. Here, these two values are $-0.0268$ and $0.1586$. But if $L'_{UICM}$ is to be treated as an unsupervised loss, normalization is also required,

\begin{equation}
L_{UICM}=\frac{L'_{UICM}-L^{min}_{UICM}}{L^{max}_{UICM}-L^{min}_{UICM}},
\end{equation}
where $L^{max}_{UICM}$ and $L^{min}_{UICM}$  represent the maximum and minimum values of $L'_{UICM}$, respectively. In this paper, the two values are $0.897177084$ and \\ $-0.0379009235$. With the normalization operation, the range of values of $L_{UICM}$ has now been fixed to between 0 and 1.

To further enhance the effect of our model output, we also added the edge-preserving loss, which was added so that the edge information in the output is not smoothed out.

\begin{eqnarray}
L_{edge}=|I_x^2-O_x^2|+|I_y^2-O_y^2|
\end{eqnarray}

In the above formula, $I_x$ and $O_x$ represent the horizontal gradient of the input image and the model output, respectively, and $I_y$ and $O_y$ represent the vertical gradient of the input image and the model output, respectively.

The goal of optimization process is to derive a model with minimum $L$, which implies the perfect pixel-level reconstruction and visual properties. The loss is defined below:

\begin{equation}
L=\lambda_1L_{pixel}-\lambda_2L_{UICM}+\lambda_3L_{edge}
\end{equation}
in which $\lambda_1$, $\lambda_2$ and $\lambda_3$ represents linear combination coefficient. In this paper, $\lambda_1=1, \lambda_2=0.001, \lambda_3=0.0001$.

\section{Experiment}

In this section, comprehensive experiments are performed on natural underwater images. The optimization method for the proposed connectionist model is Adam optimizer with initial learning rate as $0.0000125$. The structure details of our neural model are shown in Table \ref{tab:network}. The implementation of the proposed de-hazing model is conducted on Python3.5, TensorFlow1.8, GeForce GTX TITAN with 12GB RAM.

\begin{table}[h]
	\caption{Neural Model}
	\label{tab:network}       
	\begin{tabular}{lll}
		\hline\noalign{\smallskip}
        Layer type & Dimensions & Outputs \\
		\noalign{\smallskip}\hline\noalign{\smallskip}
		Convolutional & 3$\times$ 3 $\times$ 3 & 64 \\
		Convolutional & 3$\times$ 3 $\times$ 64 & 64 \\
		MaxPooling & 2 $\times$ 2 Strides 1 & - \\

		Convolutional & 3$\times$ 3 $\times$ 64 & 128 \\
		Convolutional & 3$\times$ 3 $\times$ 128 & 128 \\
		MaxPooling & 2 $\times$ 2 Strides 1 & - \\
		
		Convolutional & 3$\times$ 3 $\times$ 128 & 256 \\		
		Convolutional & 3$\times$ 3 $\times$ 256 & 256 \\		
		Convolutional & 3$\times$ 3 $\times$ 256 & 256 \\		
		Convolutional & 3$\times$ 3 $\times$ 256 & 256 \\				
		MaxPooling & 2 $\times$ 2 Strides 1 & - \\

		Convolutional & 3$\times$ 3 $\times$ 256 & 256 \\						
		Convolutional & 3$\times$ 3 $\times$ 256 & 128 \\								
		Convolutional & 3$\times$ 3 $\times$ 128 & 64 \\
		Convolutional & 3$\times$ 3 $\times$ 64 & 6 \\
		MaxPooling & 2 $\times$ 2 Strides 1 & - \\
		
		Convolutional & 3$\times$ 3 $\times$ 6 & 6 \\		
		Convolutional & 3$\times$ 3 $\times$ 6 & 6 \\
		Convolutional & 3$\times$ 3 $\times$ 6 & 6 \\
		\noalign{\smallskip}\hline
	\end{tabular}
\end{table}

\subsection{Ablation Experiments}

\begin{table}[h]
	\caption{Please write your table caption here}
	\label{tab:1}       
	\begin{tabular}{lllllll}
		\hline\noalign{\smallskip}
		& UIQM & BIQI & SSIM & MSE & PSNR & FSIM  \\
		\noalign{\smallskip}\hline\noalign{\smallskip}
		Without $L_{UICM}$ & 4.2667 & 32.7580 & 0.8285 & 0.0176 & 18.6519 & 0.8780 \\
		With $L_{UICM}$ & $\mathbf{4.6257}$ & $\mathbf{36.5955}$ & $\mathbf{0.9385}$ & $\mathbf{0.0166}$ & $\mathbf{19.0703}$ & $\mathbf{0.9372}$  \\
		\noalign{\smallskip}\hline
	\end{tabular}
\end{table}

\begin{figure*}
	\centering
	\begin{minipage}{0.18\textwidth}
		\centering
		\includegraphics[width=\textwidth]{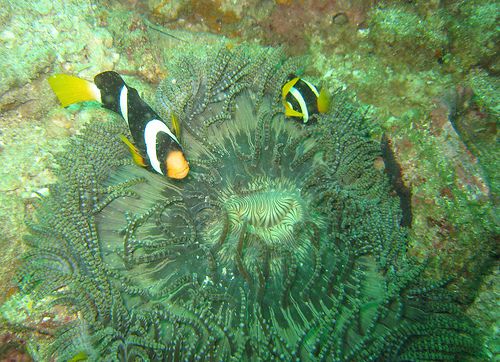}
		\subfigure{Origin}
	\end{minipage}
	\begin{minipage}{0.18\textwidth}
		\centering
		\includegraphics[width=\textwidth]{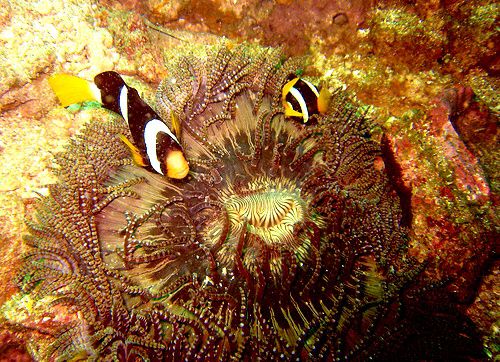}
		\subfigure{MIP}
	\end{minipage}
	\begin{minipage}{0.18\textwidth}
		\centering
		\includegraphics[width=\textwidth]{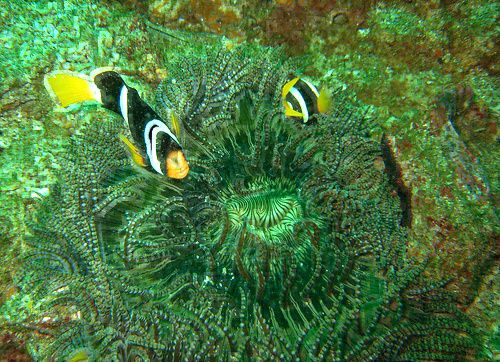}
		\subfigure{NOM}
	\end{minipage}
	\begin{minipage}{0.18\textwidth}
		\centering
		\includegraphics[width=\textwidth]{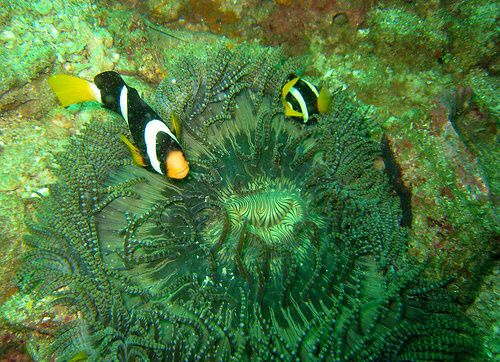}
		\subfigure{UDCP}
	\end{minipage}
	\begin{minipage}{0.18\textwidth}
		\centering
		\includegraphics[width=\textwidth]{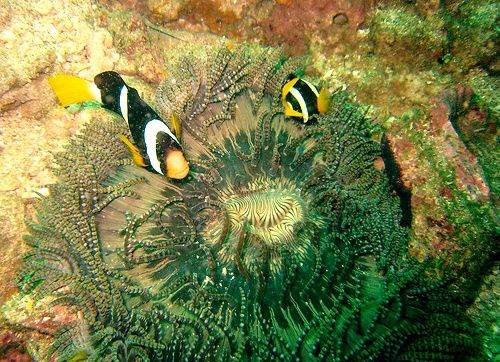}
		\subfigure{ULAP}
	\end{minipage}
	\begin{minipage}{0.18\textwidth}
		\centering
		\includegraphics[width=\textwidth]{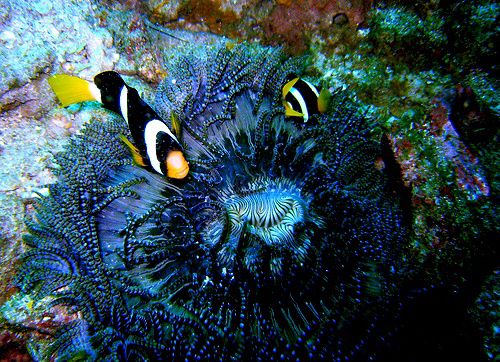}
		\subfigure{IBLA}
	\end{minipage}
	\begin{minipage}{0.18\textwidth}
		\centering
		\includegraphics[width=\textwidth]{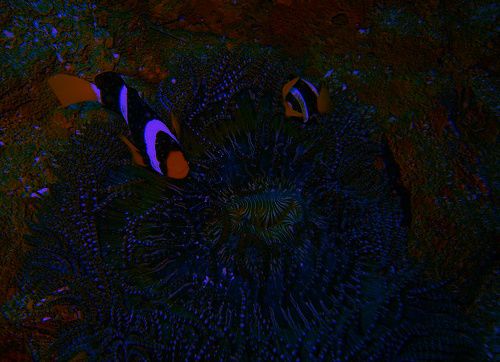}
		\subfigure{LCDCP}
	\end{minipage}
	\begin{minipage}{0.18\textwidth}
		\centering
		\includegraphics[width=\textwidth]{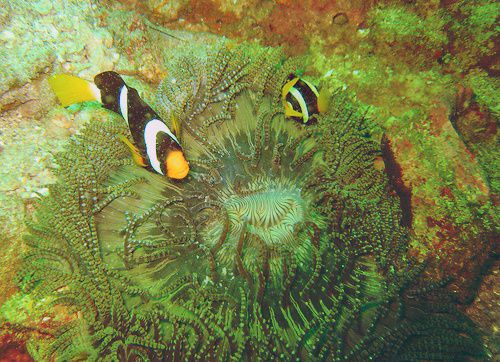}
		\subfigure{GB}
	\end{minipage}
	\begin{minipage}{0.18\textwidth}
		\centering
		\includegraphics[width=\textwidth]{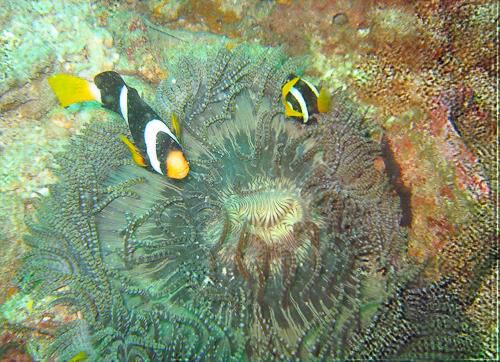}
		\subfigure{GAN}
	\end{minipage}
	\begin{minipage}{0.18\textwidth}
		\centering
		\includegraphics[width=\textwidth]{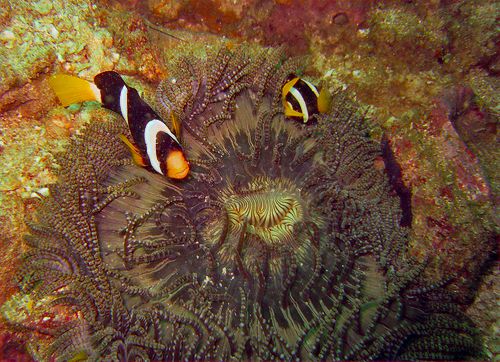}
		\subfigure{Ours}
	\end{minipage}
	
	
	\begin{minipage}{0.18\textwidth}
		\centering
		\includegraphics[width=\textwidth]{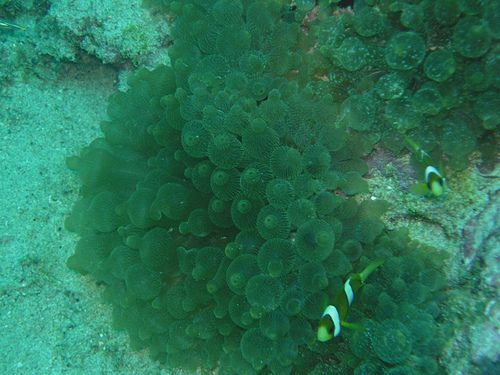}
		\subfigure{Origin}
	\end{minipage}
	\begin{minipage}{0.18\textwidth}
		\centering
		\includegraphics[width=\textwidth]{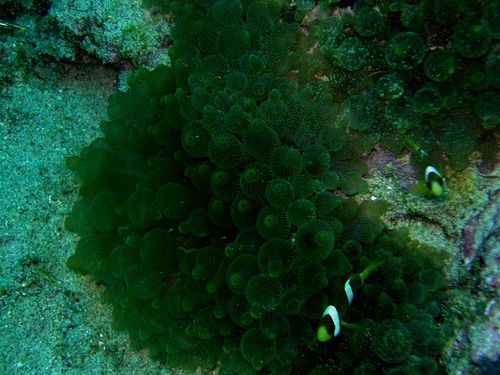}
		\subfigure{MIP}
	\end{minipage}
	\begin{minipage}{0.18\textwidth}
		\centering
		\includegraphics[width=\textwidth]{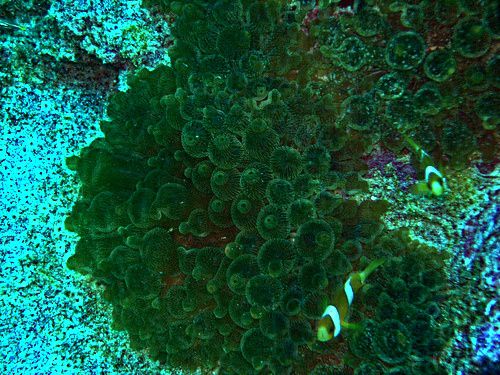}
		\subfigure{NOM}
	\end{minipage}
	\begin{minipage}{0.18\textwidth}
		\centering
		\includegraphics[width=\textwidth]{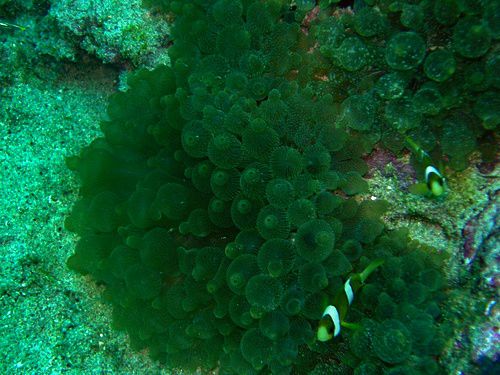}
		\subfigure{UDCP}
	\end{minipage}
	\begin{minipage}{0.18\textwidth}
		\centering
		\includegraphics[width=\textwidth]{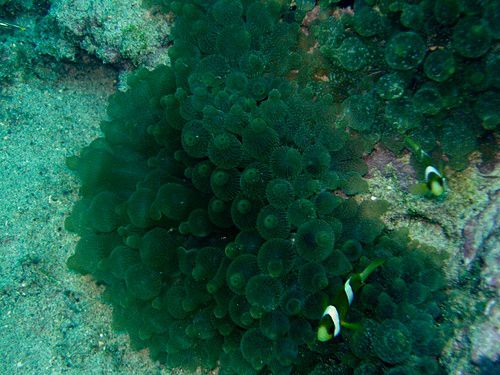}
		\subfigure{ULAP}
	\end{minipage}
	
\end{figure*}

\begin{figure*}
	\centering
	\begin{minipage}{0.18\textwidth}
		\centering
		\includegraphics[width=\textwidth]{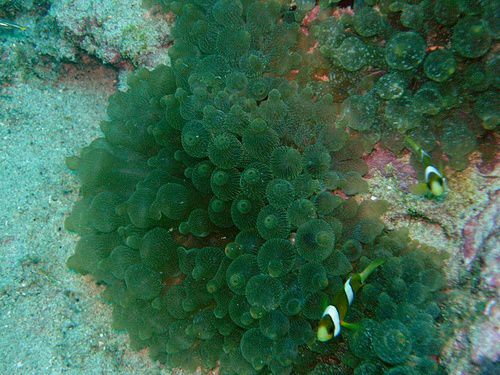}
		\subfigure{IBLA}
	\end{minipage}
	\begin{minipage}{0.18\textwidth}
		\centering
		\includegraphics[width=\textwidth]{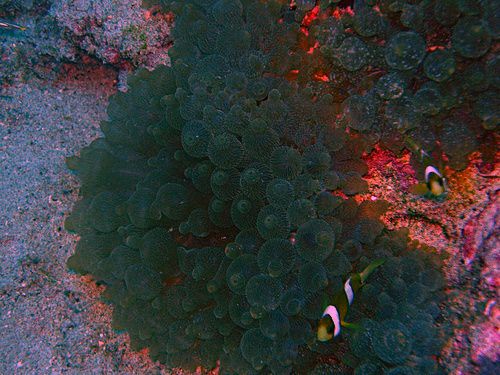}
		\subfigure{LCDCP}
	\end{minipage}
	\begin{minipage}{0.18\textwidth}
		\centering
		\includegraphics[width=\textwidth]{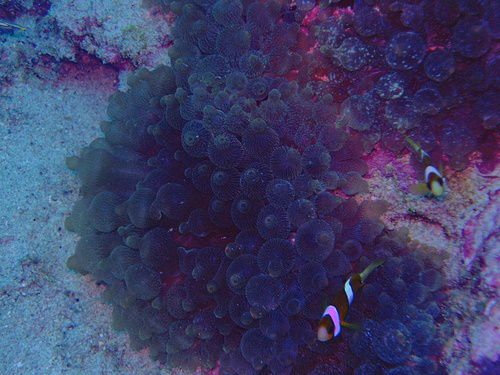}
		\subfigure{GB}
	\end{minipage}
	\begin{minipage}{0.18\textwidth}
		\centering
		\includegraphics[width=\textwidth]{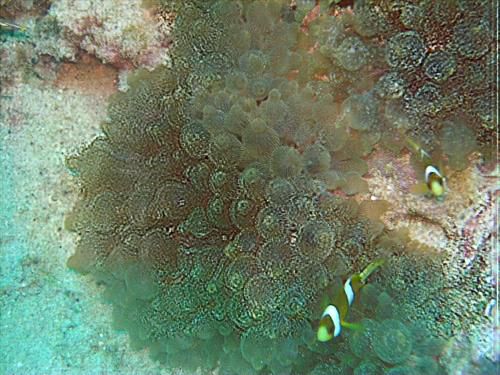}
		\subfigure{GAN}
	\end{minipage}
	\begin{minipage}{0.18\textwidth}
		\centering
		\includegraphics[width=\textwidth]{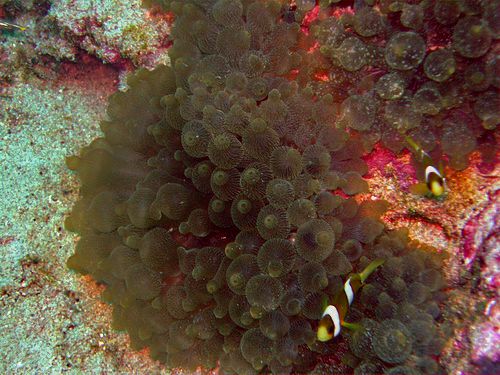}
		\subfigure{Ours}
	\end{minipage}
	
	\centering
	\caption{Experiment Result}
	\label{fig:exp}
\end{figure*}

In this experimental part, we will investigate the contribution of $L_{UICM}$ to the overall model. To this end, we compare the proposed model with the model without $L_{UICM}$. All hyper-parameters are the same both in these two models. The test data are 2000 underwater images which will be presented in following section. We report the comparison results in Table \ref{tab:1}, where UIQM\cite{UIQM}, BIQI\cite{BIQI}, SSIM\cite{SSIM}, MSE, PSRN\cite{PSNR} and FSIM\cite{FSIM} are employed to measure the quality of enhanced images. 

Through these numerical comparisons, one can note that the optimization item $L_{UICM}$ indeed improves the performance of the enhancement model with respect to image colorfulness and human perception.

\subsection{Comparison with state-of-the-art methods}

\begin{table}[h]
	\caption{Please write your table caption here}
	\label{tab:2}       
	\begin{tabular}{lllll}
		\hline\noalign{\smallskip}
        & BIQI & BRISQUE & NIQE & SSEQ\\
		\noalign{\smallskip}\hline\noalign{\smallskip}
		DCP & 34.8206 & 26.8266 & 5.2461 & 22.2326 \\
		GAN & 31.9677 & 25.0001 & $\mathbf{4.5095}$ & 26.5189  \\
		GB &  32.3765 & 25.8380 & 5.2969 & 21.6505 \\
		IBLA & 35.7431 & 27.4923 & 5.2548 & 22.1674 \\
		LDCP & 37.2837 & 31.6354 & 7.8621 & 24.2650 \\
		MIP & 36.9614 & 28.1605 & 5.6123 & 22.5312 \\
		NOM & 37.1075 & 31.7465 & 5.6184 & 24.5201 \\
		UDCP & 32.4026 & 26.9171 & 5.1882 & 22.5090 \\
		ULAP & 35.4530 & 27.0052 & 5.2952 & 21.9646 \\
		Ours & $\mathbf{40.5764}$ & $\mathbf{24.9179}$ & 4.6522 & $\mathbf{20.1862}$ \\
		\noalign{\smallskip}\hline
	\end{tabular}
\end{table}

In this part, the proposed model is compared with several popular underwater enhancement methods, namely, GB \cite{GBdehazingRCorrection}, IBLA \cite{IBLA}, LCDCP \cite{LowComplexityDCP}, MIP \cite{MIP}, NOM \cite{NewOpticalModel}, UDCP \cite{UDCP}, ULAP \cite{ULAP} and GAN \cite{GAN}. The real-world underwater images cover different situations in terms of color-casting, blur, etc. 

The visual comparisons of different underwater enhancement methods are shown in Figure \ref{fig:exp}. Through visual inspection, one can note that all methods show enhanced effects on real underwater images, but the various methods differ greatly in terms of enhancement effects. It is clear that non-deep learning image enhancement methods can cause severe color shift effects in some cases. And the method based on deep learning should be more stable in the image enhancement effect.

In addition, we also report numerical results of these methods in Table \ref{tab:2}. We employ BIQI \cite{BIQI}, BRISQUE \cite{BRISQUE}, NIQE \cite{NIQE} and SSEQ \cite{SSEQ} to measure the quality of recovered images.

This reason is that these statistical priors on the properties of images can reliably reflect the perception of the human visual system.

\section{Conclusion}

In this paper, we proposed a novel method for enhancement of underwater images. This method is based on an end-to-end neural network, which can be divided into two parts. The first one is an empty-parameter layer designed for preliminary color correction. And the following parametric layers are used for the self-adaptive refinement that is conducted by a channel-wise linear shift. The overall neural model is optimized by a joint training criterion including supervised and unsupervised objective functions. Through the comparisons on real underwater images, we show that the proposed method has competitive performance. In the future work, we will incorporate this method into other computer vision tasks, such as object detection and tracking systems.

%
%

\bibliographystyle{spmpsci}      
\bibliography{references}
%
%

\end{document}